\documentclass[conference]{IEEEtran}

\usepackage{cite}
\usepackage{url}
\usepackage[pdftex]{graphicx}
\usepackage{graphicx}
\usepackage{hyperref}

\usepackage{tikz}
\usepackage{pgfplots}
\pgfplotsset{compat=newest}
\usetikzlibrary{pgfplots.groupplots,external}
\pgfplotsset{
    discard if not/.style 2 args={
        x filter/.code={
            \edef\tempa{\thisrow{#1}}
            \edef\tempb{#2}
            \ifx\tempa\tempb
            \else
                
            \fi
        }
    }
}

\usepackage[cmex10]{amsmath}
\usepackage{amsfonts}
\usepackage{amsthm}

\usepackage{algorithm}
\usepackage{algorithmicx}
\usepackage{algpseudocode}
\usepackage[algo2e,ruled,vlined,linesnumbered]{algorithm2e}
\usepackage{xspace}
\usepackage{soul}
\usepackage{subcaption}

\usepackage{dblfloatfix}

\usepackage{multirow}

\usepackage{balance}

\graphicspath{{../}}

\allowdisplaybreaks

\IEEEoverridecommandlockouts
\IEEEpubid{\makebox[\columnwidth]{\hfill}
\hspace{\columnsep}\makebox[\columnwidth]{}}

\newcommand{\ignore}[1]{}

\newcommand{\assign}{\leftarrow}

\newcommand{\R}{\mathbb{R}}

\renewcommand{\epsilon}{\varepsilon}

\newcommand{\leadingones}{\textsc{LeadingOnes}\xspace}
\newcommand{\LO}{\textsc{Lo}\xspace}

\newcommand{\oea}{$(1 + 1)$~EA\xspace}

\newcommand{\rea}{REA\xspace}
\newcommand{\srea}{smoothREA\xspace}

\newcommand{\xold}{x_\text{old}}
\newcommand{\fold}{f_\text{old}}

\DeclareMathOperator{\old}{old}
\DeclareMathOperator{\new}{new}
\DeclareMathOperator{\mut}{flip}

\DeclareMathOperator{\Bin}{Bin}

\pgfplotscreateplotcyclelist{colorcycle}{%
red!80!black,mark=*,mark size=0.25\\
blue,mark=square*,mark size=0.25\\
cyan!80!black,mark=*,mark size=0.25\\
violet!80!black,mark=square*,mark size=0.25\\
orange,mark=*,mark size=0.25\\
green,mark=mark=square*,mark size=0.25\\
yellow, mark=*,mark size=0.25\\
}

\begin{document}

\title{Fast Re-Optimization of LeadingOnes with Frequent Changes}
\author{\IEEEauthorblockN{Anonymous Authors}}
\author{
\IEEEauthorblockN{Nina Bulanova}
\IEEEauthorblockA{
\textit{ITMO University}\\
Saint Petersburg, Russia
\\
ninokfox@gmail.com
}

\and
\IEEEauthorblockN{Arina Buzdalova}
\IEEEauthorblockA{
\textit{ITMO University}\\
Saint Petersburg, Russia
\\
abuzdalova@gmail.com
}
\and
\IEEEauthorblockN{Carola Doerr}
\IEEEauthorblockA{
\textit{Sorbonne Universit{\'e}}\\
CNRS, LIP6, Paris, France
\\
carola.doerr@lip6.fr
}
}

\maketitle

\begin{abstract}
In real-world optimization scenarios, the problem instance that we are asked to solve may change during the optimization process, e.g., when new information becomes available or when the environmental conditions change. In such situations, one could hope to achieve reasonable performance by continuing the search from the best solution found for the original problem. Likewise, one may hope that when solving several problem instances that are similar to each other, it can be beneficial to ``warm-start'' the optimization process of the second instance by the best solution found for the first. 
However, it was shown in [Doerr et al., GECCO 2019] that even when initialized with structurally good solutions, evolutionary algorithms can have a tendency to replace these good solutions by structurally worse ones, resulting in optimization times that have no advantage over the same algorithms started from scratch. Doerr et al. also proposed a diversity mechanism to overcome this problem. Their approach balances greedy search around a best-so-far solution for the current problem with search in the neighborhood around the best-found solution for the previous instance. 

In this work, we first show that the re-optimization approach suggested by Doerr et al. reaches a limit when the problem instances are prone to more frequent changes. More precisely, we show that they get stuck on the dynamic LeadingOnes problem in which the target string changes periodically.  
We then propose a modification of their algorithm which interpolates between greedy search around the previous-best and the current-best solution. We empirically evaluate our smoothed re-optimization algorithm on LeadingOnes instances with various frequencies of change and with different perturbation factors and show that it outperforms both a fully restarted (1+1) Evolutionary Algorithm and the re-optimization approach by Doerr et al.  
\end{abstract}

\sloppy{ 

\section{Introduction}
\label{sec:intro}

It is not uncommon in real-world optimization tasks that one needs to solve several problem instances of very similar type, e.g., when a routine task such as postal deliveries needs to be planned and the problem instance changes only marginally from one day to the next. Similarly, it may happen that the problem instance that we solve changes during the optimization process, e.g., because new information becomes available or some factors that determine the quality of the solution candidates change. In both situations one may hope to be able to benefit from the good solutions that were identified for the previous problem instance or from the instance before the change, respectively. Ideally, we would hope to warm-start the algorithms with the best previously found solutions, and to speed up the search compared to searching from scratch. 
While this hope may be true in most real-world optimization scenarios~\cite{dynamicSurveyBrancke}, it was shown in~\cite{DoerrDN19} that evolutionary algorithms may not always be able to benefit from such warm-starting procedure, even when the solution(s) that are transferred from one instance to the next are structurally very good. 
An extreme case for such inefficient use of information that was highlighted in~\cite{DoerrDN19} is the \leadingones problem: it was proven that even when initialized at Hamming distance one from the optimum, the number of fitness evaluations that the (1+1) Evolutionary Algorithm (EA) needs to sample the optimum for the first time can be quadratic in the problem dimension. Even worse, it was shown that this happens with constant probability when the fitness value of the solution used for warm-starting is at most $n/2$. 
Motivated by this inefficient behavior of the \oea on dynamic \leadingones, Doerr et al. introduced a re-optimization algorithm (\rea) that balances the search between sampling around the current-best solution (as done in the \oea) and around the previous-best solution. The key idea of the \rea is that not only the previous-best solution is kept in the memory, but also one point per each Hamming distance around the previous-best solution. This way, the algorithm does not suffer any more that much if the greedy search takes it on an unfavorable path away from the optimum -- the search around the previous-best solutions helps to overcome such a distraction, by re-centering the search in the interesting part of the search domain. 

\textbf{Our Results.} The theoretical analyses in~\cite{DoerrDN19} focus on re-optimization time only, i.e., on the average number of evaluations needed until a solution is found that is at least as good as the best one before the change happened. In this work, we consider a different setup, in which the fitness function is subject to frequent changes. More precisely, we (empirically) investigate how well the \oea and the \rea optimize the \leadingones problem when the optimum is changed periodically, every $\tau$ iterations, by flipping $k$ uniformly chosen bits in the target string. We observe that the \rea is indeed more efficient than the \oea in the first iterations after such a fitness function perturbation. However, the advantage of the \rea decreases and eventually vanishes as the number of iterations increases. At some point, the \oea outperforms the \rea. Based on these observations, we propose in this work a modified \rea that dynamically changes the probability to sample either in the neighborhood of the previous-best solution or around the current best solution. More precisely, the probability that our \srea selects the current-best solution as center of search slowly increases by an additive $1/(sn^2)$ term in each iteration. Here, the parameter $s$ is a ``smoothness'' parameter that determines the speed by which the \srea converges towards the \oea. We show that the \srea is more efficient than both the \oea and the \rea for broad ranges of $s$. 

\textbf{Related Work.} 
Dynamic evolutionary computation is a well-studied problem. The surveys~\cite{dynamicSurveyBrancke} and~\cite{Neumann2020dynamicSurvey} provide a large number of relevant pointers to empirical and theoretical results, respectively, so that we mention here only a few works that have appeared since then and that are most closely related to our work. 

Lengler et al. perform empirical~\cite{LenglerM20PPSN} and theoretical~\cite{LenglerR21evocop} analysis of the dynamic BinVal problem. Just like \leadingones, BinVal is a classic, well-studied function in the runtime analysis domain. It is a linear function with weights of the form $2^{n-i}$, $i=1,...,n$. In the dynamic version analyzed in the two mentioned works, the assignments of these weights to the bit positions are chosen uniformly at random in each iteration. Lengler et al. showed that an increased population size can be beneficial, in the sense that it increases the range of possible mutation rates for which the $(\mu+1)$ EA can efficiently solve the dynamic BinVal problem. It was also shown that crossover can extend the range of efficient parameter values further. Two main differences to our work exist: 
(1) the optimum remains the same in the dynamic BinVal function, whereas it is changed by $k$ bits with every change in our dynamic \leadingones problem, and 
(2) we investigate different frequencies of change, whereas Lengler et al. change the fitness function after each generation. 

Guidelines to set mutation rates and population size of \emph{non-elitist} evolutionary algorithms for the dynamic BinVal problem are provided in~\cite{LehreQ21gecco}. Extending previous work on a hand-crafted ``moving Hamming-balls'' function~\cite{DangJL17} the work also shows the benefit of non-trivial population sizes. 

A dynamic version of \leadingones was studied in~\cite{Doerr18dynamic}. In that work, perturbations may happen in each iteration. More precisely, in each iteration (independently of all previous decisions), the optimum $z$ is changed by (a) a 1-bit flip or (b) standard bit mutation with probability $p$. It is shown that if $p\le  c \ln(n)/n^2$ (in case (a)) and $p\le c \ln(n)/n^3$ (in case (b)), then the expected time until the \oea evaluates a then-optimal solution is at most $(\delta + o(1))n^{2+\delta c}$, where $\delta=(e-1)/2 \approx 0.86$. 

Related to the dynamic setting are also noisy optimization problems, and in particular models with a priori noise, in which the solution is perturbed with some probability prior to its evaluation. The main differences to our dynamic setting is that, in the noisy setting, the changes are independent in each iteration, whereas they remain fixed for the interval $\tau$ in the dynamic case studied here in this work. We nevertheless note that it has been shown that the \oea is very sensitive to such a priori noise when optimizing the \leadingones function; see~\cite{Sudholt20} for a sharp analysis and a summary of related works on the efficiency of EAs for noisy optimization.

\section{Preliminaries}
\label{sec:prelim}

We introduce the problem and the two algorithms that build the starting point for our work. Throughout this paper, we denote by $[a..b]$ the set of all integer values $r \in [a,b]$. For two bit strings $x,y \in \{0,1\}^n$ we denote by $H(x,y)=|\{ i \in [1..n] \mid x_i \neq y_i\}|$ the Hamming distance of $x$ and $y$. 

\subsection{Dynamic LeadingOnes with Frequent Changes}
\label{sec:dynLO}

The original (static) \leadingones problem is the problem of maximizing the function 
$\LO:\{0,1\}^n \to \R, x \mapsto \max\{i \in [0..n] \mid \forall j \le i: x_j=1\}$, 
which simply assigns to each bit string the number of initial ones. Given a search point $x$ with $\LO(x)=i$, it must hold that $x_1=\ldots=x_i=1$ and $x_{i+1}=0$. The only way to improve the fitness of $x$ is hence by creating a solution that is identical to $x$ in the first $i$ positions, but has the $(i+1)$-st entry flipped (i.e., replaced by $1-x_{i+1}$). 

\leadingones was introduced in~\cite{Rudolph97} to disprove a previous conjecture that for each unimodal function the number of function evaluations needed by the \oea (see Section~\ref{sec:oea}) to find an optimal solution grows sub-quadratically in the problem dimension. This was formally proven in~\cite{droste-ea}. What makes \leadingones an interesting problem for theoretical works is that it is a non-separable function, i.e., the influence that a bit has on the overall quality of the solutions may depend on the setting of other bits.

It is well understood, and formalized in the so-called \emph{unbiasedness} notion introduced in~\cite{LehreW12}, that many algorithms, and in particular most mutation-only evolutionary algorithms, are indifferent with respect to optimizing this function or any of the $\LO_{z,\sigma}$ defined as follows. For a given \emph{target string} $z \in \{0,1\}^n$ and a given permutation $\sigma$ of $[1..n]$, the \leadingones function $\LO_{z,\sigma}$ assigns to each string $x$ the function value
$\LO_{z,\sigma}(x):=\max\{i \in [0..n] \mid \forall j \le i: x_{\sigma(j)}=z_{\sigma(j)}\}$, the longest prefix (in the order prescribed by $\sigma$) of $x$ that agrees with $z$.
It is not difficult to see that $\LO_{z,\sigma}$ has a unique optimum, which is the target string $z$. In this work, we are interested in the behavior of evolutionary algorithms when this target string changes. We do not consider changes in $\sigma$. To ease the presentation of our work, we can therefore safely assume that $\sigma$ is equal to the identity. That is, all \leadingones functions appearing in our work are of the form 
$
\LO_z:\{0,1\}^n \to [0..n], x \mapsto \max\{i \in [0..n] \mid \forall j \le i: x_j=z_j\}.
$

In our \emph{dynamic variant of \leadingones with $k$-bit inversion and frequency of change $\tau$}, the target string $z$ is replaced by a new target string $z_{\new}$ after every $\tau$ steps. The Hamming distance of $z$ and $z_{\new}$ is equal to $k$, i.e., we assume that exactly $k$ bits are changed at the end of each period. We further assume that these $k$ bits are chosen uniformly at random (u.a.r.).

Note that even a 1-bit inversion can reduce the fitness of the current-best solution very drastically: in the extreme case, it decreases from $n$ to $0$ if the first bit of the target string is flipped. This example also shows that \leadingones suffers from a bad \emph{fitness-distance correlation}, in that a small fitness value does not necessarily imply a large distance to the optimum. It is this property that may lead a greedy algorithm lose track of a good solution after a target string change. More precisely, a greedy algorithm is likely to accept solutions of better fitness but larger distance from the optimum. That this situation occurs with a non-negligible probability is at the heart of the negative re-optimization result proven in~\cite{DoerrDN19}.

\subsection{The (1+1) EA with Shift Mutation}
\label{sec:oea}

The \oea algorithm (Alg.~\ref{alg:oea}) is initialized by sampling a search point $x \in \{0,1\}^n$ u.a.r. It then proceeds in rounds. In each iteration, one offspring $y$ is created from $x$ by so-called \emph{standard bit mutation}. Standard bit mutation creates a copy from $x$ and then changes each bit with probability $p$, independently of the status of all other bits. The \emph{offspring} $y$ replaces $x$ if it is at least as good, i.e., if $f(y) \ge f(x)$ holds. The algorithm proceeds this way until a termination criterion is met. In our work, we run all algorithms for a certain number of iterations. 

The standard setting for the \emph{mutation rate} $p$ is $1/n$. With this choice, one bit is changed on average, which can be easily verified by considering that the number of bits that change are binomially distributed. However, for $p=1/n$ it happens with probability $(1-1/n)^n \approx 1/e \approx 36.8\%$ that none of the bits is flipped. We follow a suggestion made in~\cite{practice-aware} and apply the so-called \emph{shift} operator: when none of the bits is changed, we flip a randomly selected bit. Put differently, our mutation operator first selects a mutation strength $\ell \in [1..n]$ by sampling from $\Bin_{0 \rightarrow 1}(n,p)$, which assigns probability $\Bin(n,p)(k)=\binom{n}{k}p^k(1-p)^{n-k}$ to each $k \in [2..n]$, and probability $\Bin(n,p)(0)+\Bin(n,p)(1)=(1-p)^n+np(1-p)^{n-1}$ to $k=1$. 
The offspring $y$ is then created from $x$ by choosing $\ell$ pairwise different positions $i_1,\ldots,i_{\ell} \in [1..n]$ and by setting $y_j=1-x_j$ for $j \in \{i_1,\ldots,i_{\ell}\}$ and by setting $y_j=x_j$ otherwise. This procedure is implemented by the $\mut_{\ell}$ operator mentioned in line~\ref{line:mutoea} of Alg.~\ref{alg:oea}.

 \begin{algorithm2e}[t]%
	\textbf{Initialization:}
	Sample $x \in \{0,1\}^{n}$ u.a.r. and evaluate $f(x)$\;
  \textbf{Optimization:}
	\For{$t=0,1,2,\ldots$}{
		Sample $\ell$ from $\Bin_{0 \rightarrow 1}(n,p)$, sample
		$y \assign \mut_{\ell}(x)$ and evaluate $f(y)$\label{line:mutoea}\;
		\lIf{$f(y)\ge f(x)$}{$x \assign y$}
	}
\caption{The \oea with shift mutation and mutation rate $0<p<1$ maximizing a function $f:\{0,1\}^n \rightarrow \R$.}
\label{alg:oea}
\end{algorithm2e}

\subsection{The Original REA Algorithm}
\label{sec:REA}

It was shown in~\cite{DoerrDN19} that the \oea can lose track of the good solutions when a $k$-bit inversion happens on \leadingones, in the sense that the average time required to regain a solution of previous fitness can be almost as long as when started from scratch, and this even when only a single bit changed. Doerr et al. therefore suggested a \emph{Re-optimization EA (REA)}, which works as follows (see also Alg.~\ref{alg:old-rea}).

\begin{algorithm2e}[t]%
\textbf{Input:} Solution $\xold$\;
\textbf{Initialization:} $x^{0},x^{*} \assign \xold$\; \label{line:init-s} 
\Indp
    \lFor{$i=1,2,\ldots,\gamma+1$}{$x^i \assign \text{undefined}$, $f^{i} \assign -\infty$} \label{line:init-e}
 \Indm
\textbf{Optimization:}
\For{$t=0,1,2,\ldots$}{
    Select parent $x$ by choosing $x^{*}$ with probability $1/2$ and uniformly at random from $\{x^i \mid i \in [0..\gamma+1]\} \setminus \{x^{*}\}$ otherwise\label{line:sel}\;
    Sample $\ell$ from $\Bin_{0 \rightarrow 1}(n,p)$, sample
		$y \assign \mut_{\ell}(x)$ and evaluate $f(y)$\label{line:mutreaold}\;
    \lIf{$f(y) \ge f(x^*)$}{\label{line:selectionrea}$x^{*} \assign y$} \label{line:update-s}
    $i \assign \min\{H(y,\xold),\gamma+1\}$\; 
    \lIf{$f(y)\ge f^i$}{$x^i \assign y$, $f^i \assign f(y)$} \label{line:update-e}
}
\caption{The original \rea for the re-optimization (here: maximization) of a function $f:\{0,1\}^n \to \R$, which emerged from the function $\fold$ by a dynamic change.}
\label{alg:old-rea}
\end{algorithm2e}

The \rea takes as an input the individual $x_\text{old}$ with the best fitness $f_\text{old}(x_\text{old})$ achieved during the previous period before the fitness function $f_\text{old}$ was changed to $f$. 
Then a special set of individuals is initialized in lines~\ref{line:init-s}-\ref{line:init-e} and updated in lines~\ref{line:update-s}-\ref{line:update-e}. It consists of individuals which are at certain Hamming distances from $x_{old}$. More precisely, one individual $x^i$ per each Hamming distance from 0 to $\gamma$ is stored, and there is also one additional individual $x^{\gamma+1}$ which can be at any distance greater than $\gamma$. The parameter $\gamma$ needs to be chosen by the user, and is usually an upper bound for the estimated difference between the old and the new optimum. In our work, we will assume that the number $k$ of bit inversions is known, and set $\gamma=k$. This is the optimal choice of $\gamma$.   

In each iteration, the \rea selects one \emph{parent} $x$ from the current population, and creates one offspring $y$ using the same shift mutation operator $\mut_{\ell}$ as the \oea. Two greedy selection steps follow: the current-best solution $x^*$ is replaced by $y$ if $f(y)$ is at least as large as $f(x^*)$. In addition, $y$ replaces the best-so-far solution $x^i$, $i=\min\{H(x,y),\gamma+1\}$, if $f(y) \ge f^i$.
The parent selection is done as follows (line~\ref{line:sel}): with probability $1/2$ the current-best solution $x^*$ is chosen. Otherwise, the algorithm selects u.a.r. among all other members of the population $\{x^i \mid i \in [0..\gamma+1]\}\setminus\{x^*\}$. 

Thus, in each iteration, the \rea behaves like the \oea with probability $1/2$ and it decides to search around a point with possibly small fitness value otherwise. It was shown in~\cite{DoerrDN19} that the \rea needs at most $\min\{2e(\gamma+1)k n, 2en^2\}$ iterations to find again a solution of fitness at least as large as the best fitness found before the $k$-bit inversion, provided that $\gamma \ge k-1$ holds. Note that this result was for the \rea version that uses standard bit mutation and not the shifted mutation, but it is not difficult to see that the shift mutation changes only the constants in the bound (this can be derived similarly as in~\cite{practice-aware}). We omit the details, as we are mainly interested in the global picture.   

\section{Modified and Smooth REA for Frequent Changes of the Fitness Function}
\label{sec:ourREA}

While the main focus in~\cite{DoerrDN19} is on the re-optimization time, i.e., on the time needed to find again a search point that is at least as good as the best one that was evaluated before the dynamic perturbation of the fitness function, we are interested in this work in situations in which the $k$-bit inversion happens with a certain frequency $\tau$. We therefore need to extend the \rea by switching off the re-optimization part when a search point of fitness at least $f_{\old}(x_{\old})$ is found. When this is the case, i.e., when the \rea has found a solution that is at least as good as the best one found for the function before the $k$-bit inversion, the algorithm cannot benefit any more from the population $\{x^ i \mid i \in [1..\gamma+1]\}$. In this situation, we therefore let the \rea continue as a \oea (line~\ref{line:oeapart} of Alg.~\ref{alg:new-rea}). For reasons of space, we do not provide the full pseudo-code of the REA adjusted to our setting, but it is the same as Alg.~\ref{alg:new-rea} with line~\ref{line:mod-sel} replaced by line~\ref{line:sel} from Alg.~\ref{alg:old-rea}. Despite the extension from the original \rea proposed in~\cite{DoerrDN19}, we continue to refer to this algorithm as the \rea (there is no risk of confusion since we only consider this extended version in this work). 

During preliminary experiments, it was repeatedly noticed that the \rea has a clear advantage over the \oea in optimization speed immediately after the fitness function perturbation, but that it begins to fall behind the \oea after some time. This behavior is also visible in the plots presented in Section~\ref{sec:results}, e.g., Figure~\ref{fig:3bits}. 
It inspired us to gradually change the probability by which the \rea resembles the \oea. We use a simple linear probability adjustment: in iteration $t$ after the last change, the probability to select the best-so-far solution $x^*$ is set to $\min\{1,t/(sn^2)\}$, where $s>0$ is a hyper-parameter of our modified \rea. We call $s$ the \emph{smoothness parameter.} 
It determines the speed by which the modified REA converges from uniform selection among the points that are not the current-best\footnote{One may wonder if it would not be better to select uniformly among all points in the population, but we did not find evidence in our preliminary experiments that such a strategy would be advantageous over the one adopted in Alg.~\ref{alg:new-rea}.} towards greedy selection of the latter. The smaller the value of $s$, the faster the modified \rea converges to the \oea. Note that we have chosen the normalization by $n^2$ since the optimization time of the \oea on \leadingones is $\Omega(n^2)$~\cite{droste-ea}. We refer to this modified \rea as the \srea.

\begin{algorithm2e}[t]
	\textbf{Input:} Smoothness parameter $s>0$\\
	\textbf{Initialization:}
	$\text{re-optimization flag } r \gets \text{false}$; \\
	Sample $x^* \in \{0,1\}^{n}$ u.a.r. and evaluate $f(x^*)$  \\
	$x_\text{old}  \gets x^*$\\
	$f_\text{best} \gets f(x^*)$\\
	\textbf{Optimization:}
	\For{$t=0,1,2,\ldots$}{
		\If{$t \mod \tau = 0 \wedge t \neq 0$}
		{
			$x_\text{old} \gets x^{*}$\;
			$f_\text{best} \gets f(x_\text{old})$\;
			$\text{FitnessFunctionPerturbation}(f)$ \\ $k$-bit inversion of target string\;
			$r \gets \text{true}$\;
			$x^{0} \gets x_\text{old}$\;
			\lFor{$i=1,2,\ldots,\gamma+1$}{$x^i \gets \text{undefined}$, $f^{i} \gets -\infty$}

		}
		\uIf{$r$}
		{
			Select parent $x$ by choosing $x^{*}$ with probability 
			$\min\{1,t/(sn^2)\}$  
			and uniformly at random from $\{x^i \mid i \in [0..\gamma+1]\} \setminus \{x^{*}\}$ otherwise\label{line:mod-sel}\;

		}
		\Else {
			$x \gets x^{*}$\label{line:oeapart}\;
		}

		 Sample $\ell$ from $\Bin_{0 \rightarrow 1}(n,p)$, sample
		$y \assign \mut_{\ell}(x)$ and evaluate $f(y)$\label{line:mutreanew}\;
		\lIf{$f(y) \ge f(x^*)$}{\label{line:selection}$x^{*} \assign y$}

		\If{$r$}
		{
			$i \assign \min\{H(y,\xold),\gamma+1\}$\;
   			 \lIf{$f(y)\ge f^i$}{\label{line:update}$x^i \assign y$, $f^i \assign f(y)$}
			\If{$f(x^{*}) \ge f_\text{best}$}
			{
				$r \gets \text{false}$\;

			}
		}
	}
	\caption{The \srea for maximizing a dynamic function that changes every $\tau$ iterations. 
	The (extended) \rea is identical to this algorithm after replacing line~\ref{line:mod-sel} with line~\ref{line:sel} of Alg.~\ref{alg:old-rea}.}
	\label{alg:new-rea}
\end{algorithm2e}

\section{Results}
\label{sec:results}

\begin{figure*}[ht!]
	\centering
	\includegraphics[width=0.7\textwidth]{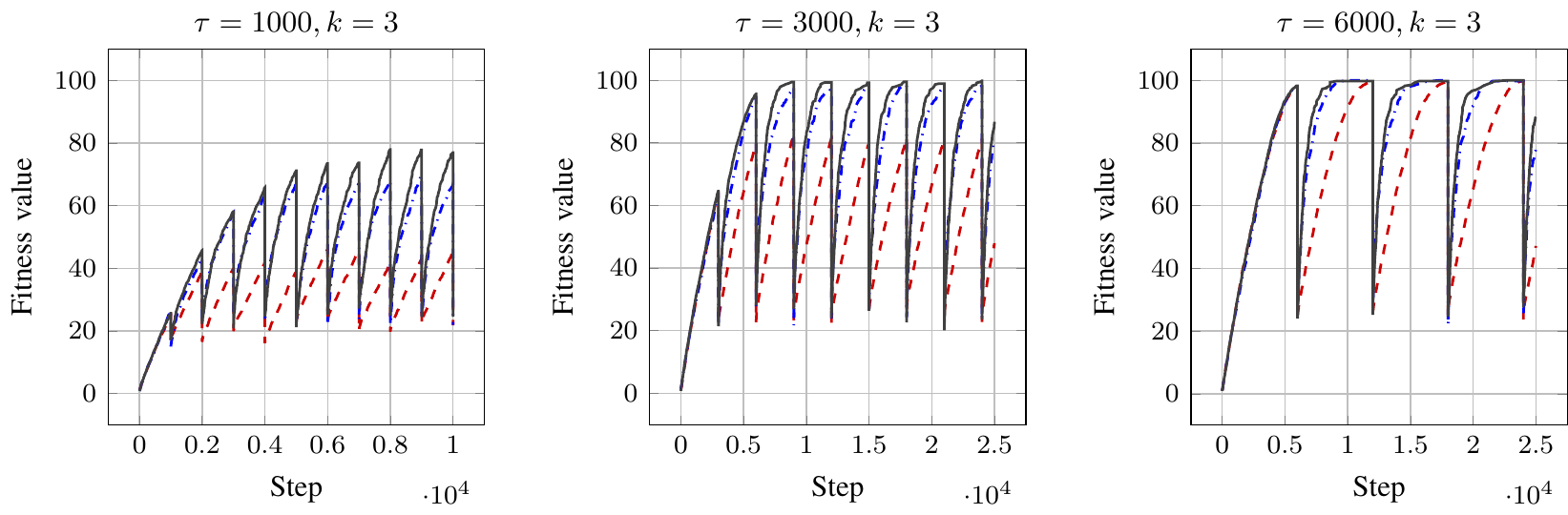}
	\includegraphics[width=0.7\textwidth]{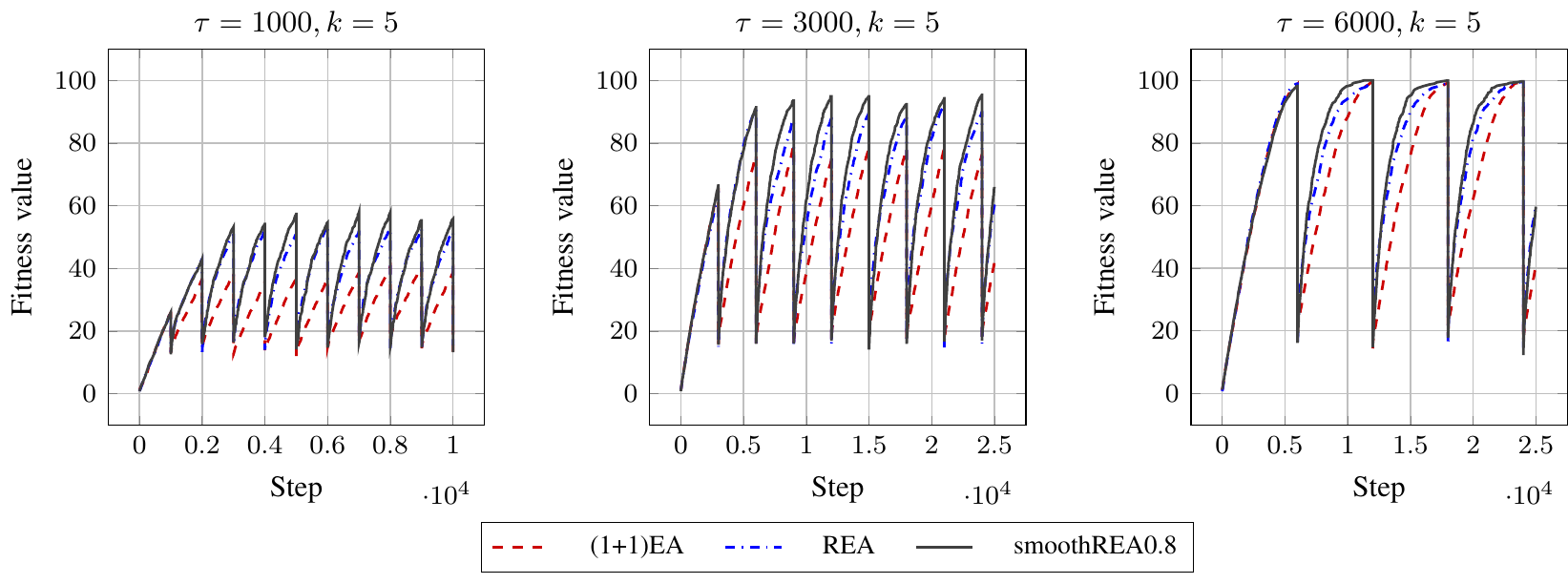}
	\includegraphics[width=0.7\textwidth]{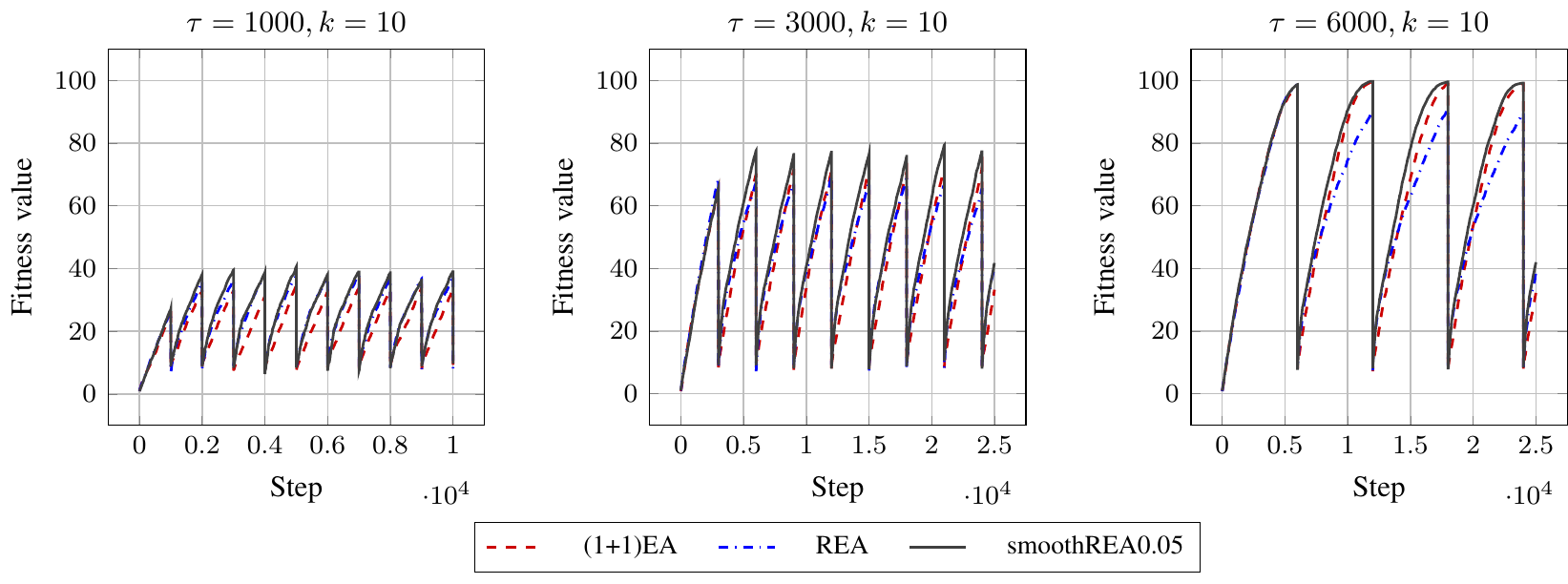}
	\caption{Average function values of the \oea, \rea, and \srea for different instances of the dynamic \leadingones problem with $k$-bit inversion and problem size $n=100$. 
	Results are grouped by $k$ in the rows (top: $k=3$, middle: $k=5$, bottom: $k=10$) and for different frequencies of change in the columns. The smoothness parameter $s$ is set to $0.8$ for $k \in \{3,5\}$ and we show results for $s=0.05$ for $k=10$.}
	\label{fig:3bits}
  \end{figure*}

\begin{figure*}[h!]
	\centering
	\includegraphics[width=0.75\textwidth]{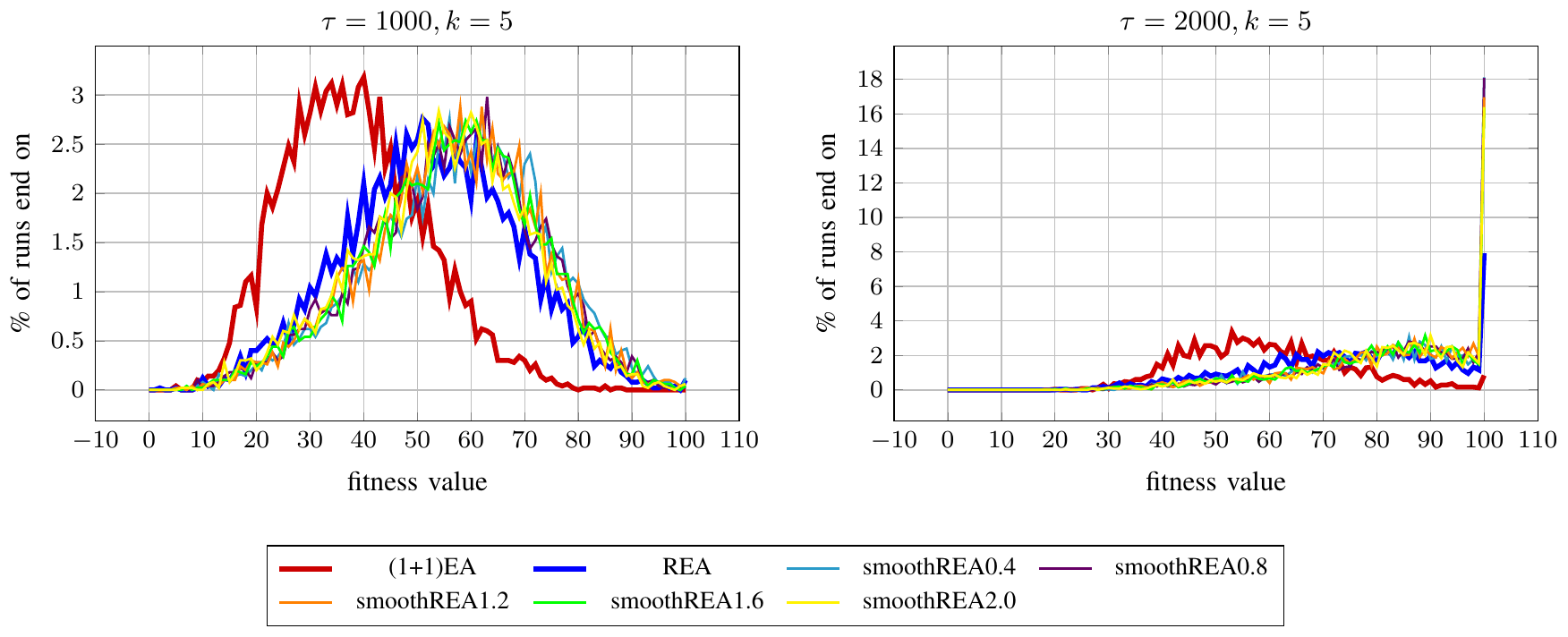}
	\caption{Histograms of best fitness values found in each period between two fitness function perturbations, for dynamic \leadingones with frequency of change $\tau = 1000$ (left) and $\tau = 2000$ (right), perturbation strength $k=5$, problem dimension $n=100$, and 100 independent runs of the algorithms.}
	\label{fig:histo}
  \end{figure*} 

\begin{figure*}[ht]
	\centering
	\includegraphics[width=0.7\textwidth]{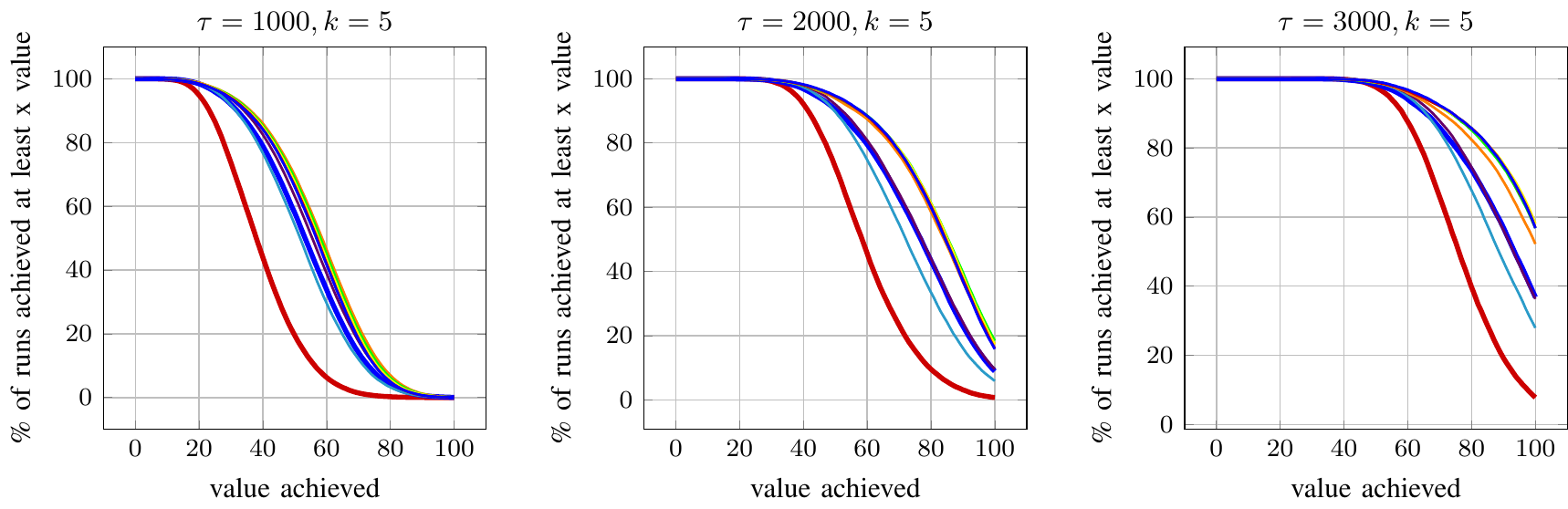}
	\includegraphics[width=0.7\textwidth]{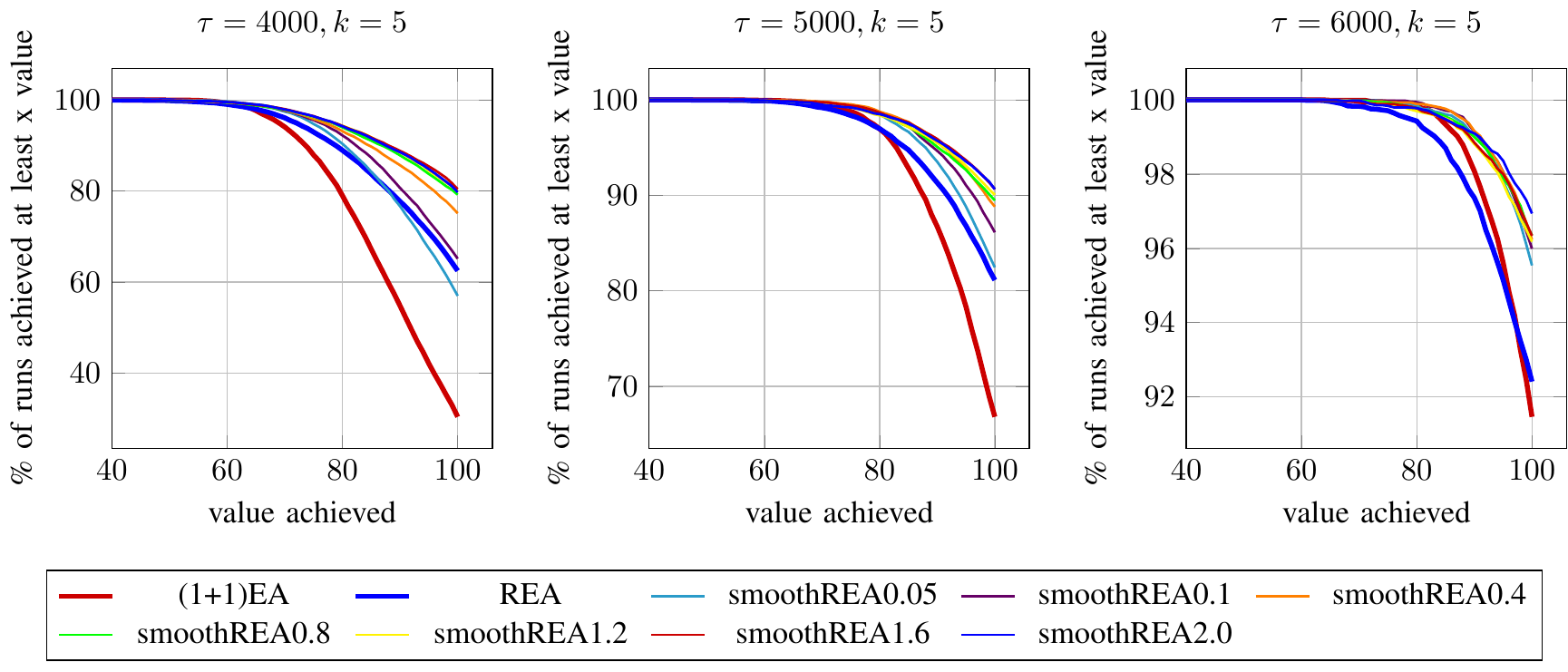}
	\caption{Cumulative plots of the values plotted in Figure~\ref{fig:histo}, for 1000 runs. Each point in the plot shows the fraction of values that is at least as large as the value indicated by the x-axis. All results are for the dynamic \leadingones problem with perturbation strength $k=5$, dimension $n=100$, and for frequencies of change ranging from $\tau=1000$ to $\tau=6000$.}
	\label{fig:cumulative}
  \end{figure*}

\begin{figure*}
	\centering
	\includegraphics[width=0.8\textwidth]{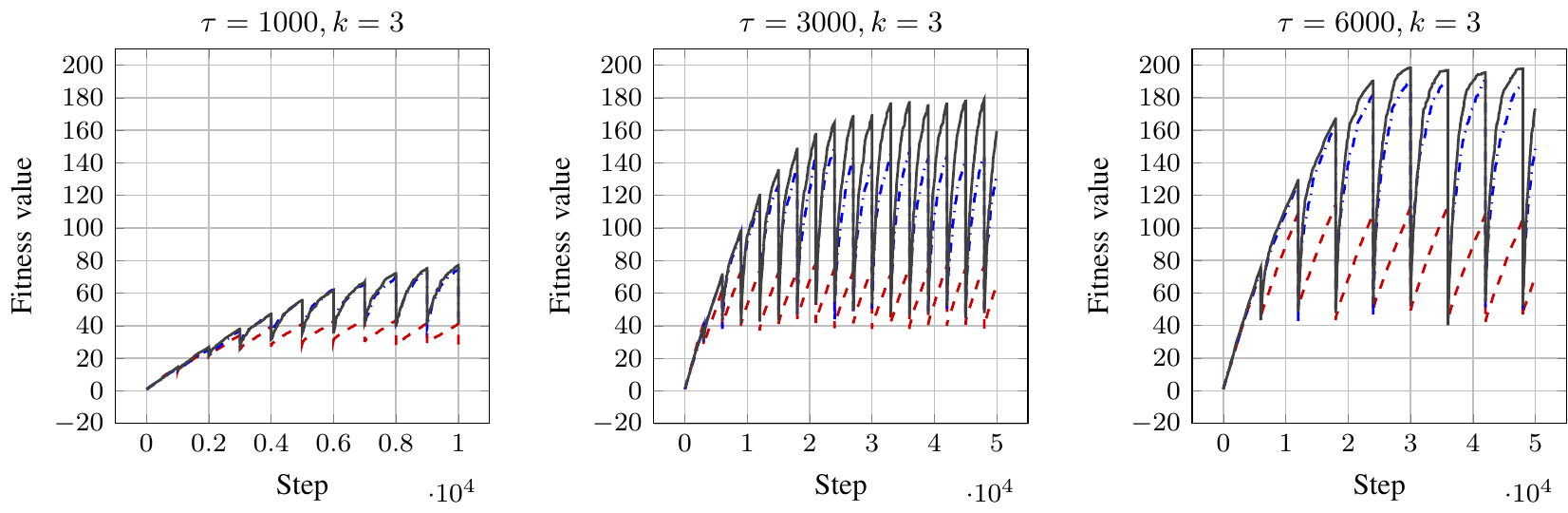}
	\includegraphics[width=0.8\textwidth]{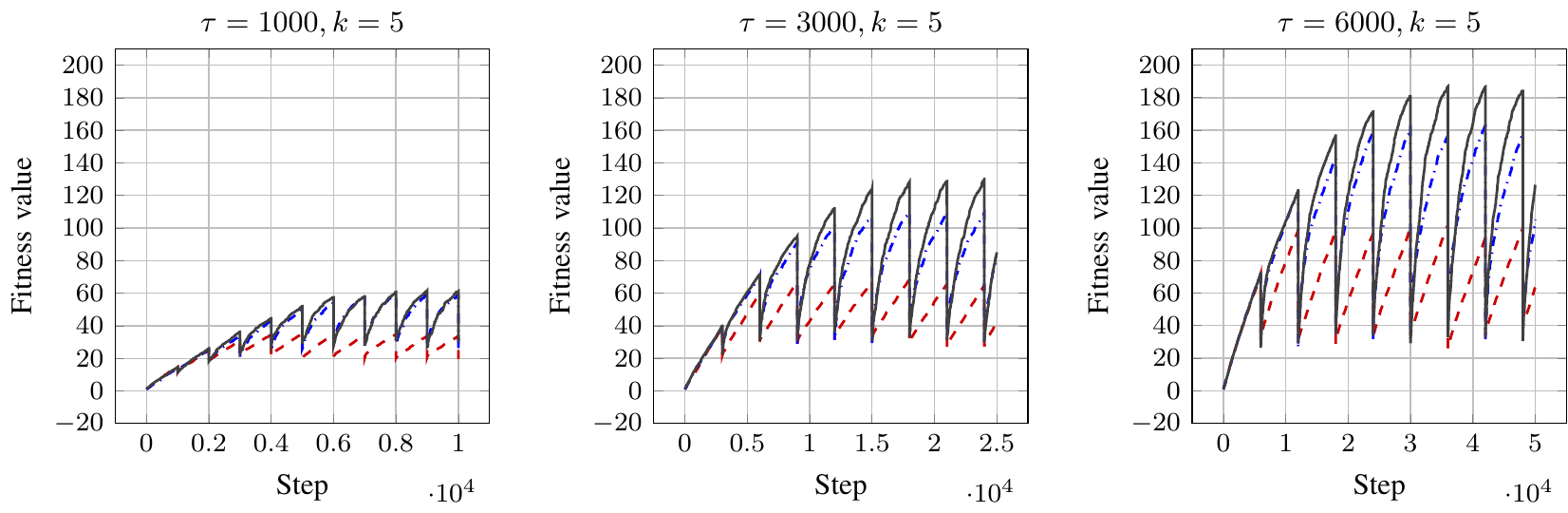}
	\includegraphics[width=0.8\textwidth]{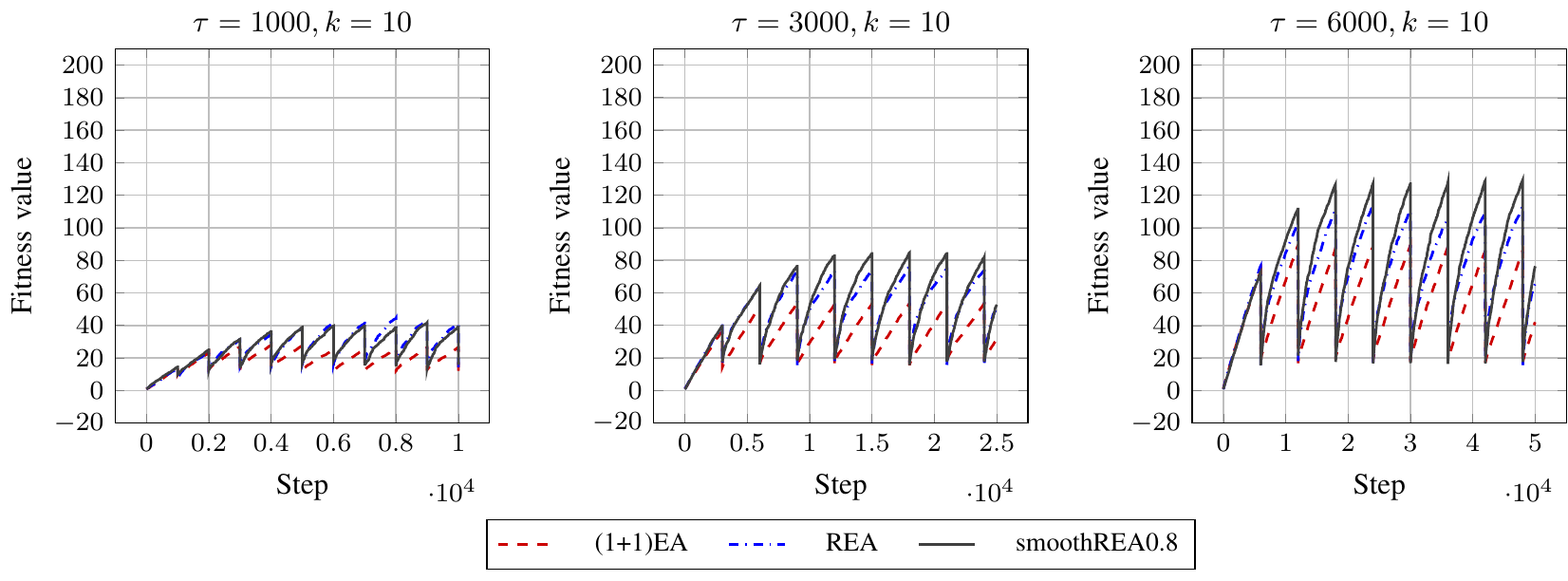}
	\caption{Average function values of the \oea, \rea, and \srea for different instances of the dynamic \leadingones problem with $k$-bit inversion and problem size $n=200$. 
	Results are grouped by $k$ in the rows (top: $k=3$, middle: $k=5$, bottom: $k=10$) and for different frequencies of change in the columns. The smoothness parameter $s$ is set to $0.8$ for all settings.}
	\label{fig:3bits200}
  \end{figure*}

\subsection{Experimental Setup}
\label{sec:setup}

We compare the \oea, the \rea, and the \srea on the following dynamic \leadingones instances and with the following settings:  
\begin{itemize}
\item dimension of the problem: $n \in \{100, 200\}$, 
\item frequency of fitness function perturbations $\tau \in \{500\} \cup \{1000i \mid i \in [1..10]\}$,
\item strength of the perturbation, i.e., number of bits flipped in every fitness function perturbation: $k \in \{3, 5, 10\}$,
\item size of the Hamming set: as motivated in Section~\ref{sec:REA} we use the optimal choice $\gamma=k$,
\item smoothness parameter\\ $s \in \{0.05, 0.1, 0.4, 0.8, 1.2, 1.6, 2.0\},$
\item total budget of function evaluations: 50\,000 (the visible part on the plots may be less),
\item results are averaged based on 100 runs for the performance plots (Figures~\ref{fig:3bits} and~\ref{fig:3bits200}) and the histograms in Figure~\ref{fig:histo}. To increase the smoothness of the curves, they are based on 1000 runs for the cumulative plots shown in Figure~\ref{fig:cumulative}.
 \end{itemize}

Although we do not explicitly report additional statistical measures, such as deviation, they can be assessed using the histograms  in Figure~\ref{fig:histo}.

The source code used to perform this experimental study may be found at \url{https://github.com/Ninokfox/REA}.

\subsection{Overview of Results, $n=100$}
\label{sec:resultsn100}

Figure~\ref{fig:3bits} shows the average quality of the best-so-far points obtained by the three algorithms in dependence of the number of solutions that have been evaluated, for different frequencies of change ($\tau \in \{1000, 3000, 6000\}$) and for different perturbation strengths ($k \in \{3,5,10\}$). For the \srea we plot here only the results for two selected values, $s=0.8$ for $k\in \{3,5\}$ and $s=0.05$ for $k=10$. These values were selected because they had decent performance. Results for other smoothness values will be discussed below. 

Overall, \rea performs better than the \oea for most of the considered cases, but not for the setting with $k=10$ and $\tau \ge 3000$. This is in line with the theoretical result mentioned at the end of Section~\ref{sec:REA}: the larger $k$, the less likely the \rea is to select a good starting point, hence its reduced performance. 
Our proposed \srea outperforms the other two algorithms in all cases, but the advantage is almost negligible for $k=10$. 

We also observe that for all three algorithms the average best fitness value stagnates when the frequency of change is too large ($\tau=1000$ for $k=3$, $\tau \le 3000$ for $k=5$ and $k=10$). The value at which the algorithms stagnate is typically lowest for the \oea and largest for the \srea.

In order to trace the influence of the smoothness parameter $s$ on the efficiency of the \srea, 
we consider the fitness values of best-found solutions at the last iteration before a perturbation of the fitness function happens. 
We plot these fitness values in histograms in Figure~\ref{fig:histo}. More precisely, we plot the results for the \oea, the \rea, and for the \srea  algorithms with smoothness values $s \in \{0.4, 0.8, 1.2, 1.6, 2.0\}$, for the case when $k=5$ bits are flipped in the target string. The plot on the left shows the distribution of values for $\tau=1000$ and the one on the right for $\tau=2000$.  Each plot is based on $100$ (number of runs) times $50\,000/\tau$ fitness values, i.e, the plot on the left shows the distribution for $5\,000$ values, and the one on the right for $2\,500$ values.

For $\tau=1000$, the distributions look like normally distributed ones for each of the seven algorithms. The mean of the values found by the \oea are smaller than those found by the \rea and the \srea algorithms. The difference between the original \rea and the family of \srea algorithms is noticeable as well, but is much smaller than the difference to the \oea. 

For both frequencies of changes plotted in Figure~\ref{fig:histo} the distribution of the values look very similar for the different smoothness parameters $s$. 

For $\tau = 2000$, the difference between the \rea and the family of \srea algorithms is more pronounced, especially when considering the number of runs that hit the optimum (fitness equal to 100). Since this effect is not very well visible in this plot, we show a cumulative version of the plots in Figure~\ref{fig:cumulative}, for different frequencies of change. More precisely, we plot in Figure~\ref{fig:cumulative} which percentage of the values (y-axis) are at least as large as the value indicated on the x-axis.

When the perturbations are less frequent (i.e., when $\tau$ increases), the number of periods in which the algorithms reach the optimum increases noticeably. The \oea (red line) is clearly worse than the \rea (thick blue line) and all the \srea variants (thin lines), with the exception of the case $\tau=6000$ where the \rea achieves the optimum more often than the \oea, but the \oea has a larger fraction of periods in which it reaches fitness values up to 95. 

We also observe that the \rea and the \srea algorithms are very similar for high frequencies of change, especially for $\tau=1000$. But the difference between them increases with increasing $\tau$.

Overall, the visualization in Figure~\ref{fig:cumulative} clearly indicates that the dispersion of the fitness values reached at the end of each period is larger for the \rea than for the \oea, which explains the poor average performance plotted in Figure~\ref{fig:3bits}. 

Comparing the different \srea variants, we observe that for high frequency of change $\tau = 1000$ the algorithms with $s=0.4$ and $s=0.8$ are the most efficient, for the frequency $\tau = 2000$ smoothness values $s=0.8$ and $1.2$ are the best choices, for $\tau = 4000$ one should pick $s = 1.6$ and $2.0$ and so on. That is, the smoothness parameter $s$ should increase with increasing periods between two fitness function perturbations. Put differently, the less frequent the change occurs, the less pronounced the benefit of the \rea mechanism, and the faster one should change to the greedy selection (i.e., to the \oea).

\subsection{Overview of Results, $n=200$}
\label{sec:resultsn200}

Figure~\ref{fig:3bits200} shows the average best-so-far fitness that the three algorithms achieved on different instances of the dynamic \leadingones problem in dimension $n=200$. For a better comparability, the frequencies of change and perturbation strengths are identical to those plotted for the case $n=100$ in Figure~\ref{fig:3bits}.  

As expected, the average best fitness ever found is smaller for the $200$-dimensional problem compared to the 100-dimensional one. For all nine settings, none of the three algorithms is able to reliably locate the optimal solution. 
Naturally, if the frequency and strength of perturbation stay the same, the percentage of changed bits relative to the length of the individual is halved in the $n=200$ setting compared to $n=100$. This makes the optimization for $k = 10$ and $n = 200$ similar to the optimization for $k = 5$ and $n = 100$. 
The same smoothness parameter turned out to be efficient in both cases. 
We also observe that the \rea outperforms the \oea in all cases, i.e., even for all three cases with perturbation strengths $k=10$. 

As mentioned in Section~\ref{sec:setup}, we have tested various smoothness values. While $s=0.8$ is among the best for most experiments, smaller values become beneficial when the frequency of change increases. For example, for $\tau=1000$ and $k=3$, the \srea with $s=0.05$ is more efficient than the one with $s=0.8$, but the difference in performance is not very large.

\section{Conclusion}
\label{sec:concl}

We have analyzed the performance of the \rea proposed in~\cite{DoerrDN19} in the context of high frequencies of change. We first modified the \rea for this setting, by switching off the \rea elements when a solution is found that is at least as good as the best solution that was found for the fitness function before the last perturbation. In this case, the modified \rea continues as a \oea. We then observed that this modified \rea is efficient only for the first steps after the fitness function perturbation, but falls behind the efficiency of the \oea after a certain number of steps. We have therefore proposed a family of \srea algorithms, which interpolate the \rea with the \oea. We show that this version improves over the original \rea for a broad range of smoothness parameters~$s>0$. 

In contrast to the work presented in~\cite{DoerrDN19}, our analysis is purely empirical. Even though our algorithms use the shift mutation operator suggested in~\cite{practice-aware}, it should not be too difficult to compute the exact point at which one should ideally switch off the \rea elements and continue with the \oea. Likewise, it should not be too difficult to compute the optimal smoothness parameter~$s$ in dependence of $n$ and $k$. 

All algorithms investigated in this work struggle to find solutions that are better than a certain threshold that depends on $n$, $k$, and $\tau$. There is hence room for further improvements of the re-optimization strategy. 

Leaving the world of \leadingones instances, we plan on investigating the basic working principles of the \rea on dynamic combinatorial problems such as dynamic scheduling or dynamic routing problems~\cite{xu2021, diao2019}. 

\section*{Acknowledgments} 

This work was supported by the Analytical Center for the Government of the Russian Federation (IGK 000000D730321P5Q0002), agreement No. 70-2021-00141.
We thank Danil Shkarupin for initial explorations.


\balance
}


\end{document}